\documentclass[journal]{vgtc}                     

\onlineid{1620}

\vgtccategory{Research}
\usepackage{paralist}
\usepackage{enumitem}
\usepackage{xcolor}
\usepackage{xspace}

\newcommand{\tool}{\textit{VeriLLMed}\xspace}

\newcommand{\revise}[1]{#1}
\newcommand{\cl}[1]{}
\newcommand{\cll}[1]{}

\title{\tool: Interactive Visual Debugging of Medical Large Language Models with Knowledge Graphs}

\author{%
  Yurui Xiang,
  Xingyi Mao,
  Rui Sheng,
  Zixin Chen,
  Zelin Zang,
  Yuyang Wu,\\
  Haipeng Zeng,
  Huamin Qu,
  Yushi Sun\textsuperscript{*}, and
  Yanna Lin\textsuperscript{*}
}

\authorfooter{
  \item
    * Corresponding authors: Yushi Sun and Yanna Lin.
}

\abstract{%
Large language models (LLMs) show promise in medical diagnosis, but real-world deployment remains challenging due to high-stakes clinical decisions and imperfect reasoning reliability. 
As a result, careful inspection of model behavior is essential for assessing whether diagnostic reasoning is reliable and clinically grounded.
However, debugging medical LLMs remains difficult.
First, developers often lack sufficient medical domain expertise to interpret model errors in clinically meaningful terms.
Second, models can fail across a large and diverse set of instances involving different input types, tasks, and reasoning steps, making it challenging for developers to prioritize which errors deserve focused inspection.
Third, developers struggle to identify recurring error patterns across cases, as existing debugging practices are largely instance-centric and rely on manual inspection of isolated failures.
To address these challenges, we present \tool, a visual analytics system that integrates external biomedical knowledge to audit and debug medical LLM diagnostic reasoning.
\tool transforms model outputs into comparable reasoning paths, constructs knowledge graph-grounded reference paths, and identifies three recurring classes of diagnosis errors: missing errors, relation errors, and branch errors.
Case studies and expert evaluation demonstrate that \tool helps developers identify clinically implausible reasoning and generate actionable insights that can inform the improvement of medical LLMs.
}


\teaser{
  \centering
  \includegraphics[width=0.92\linewidth]{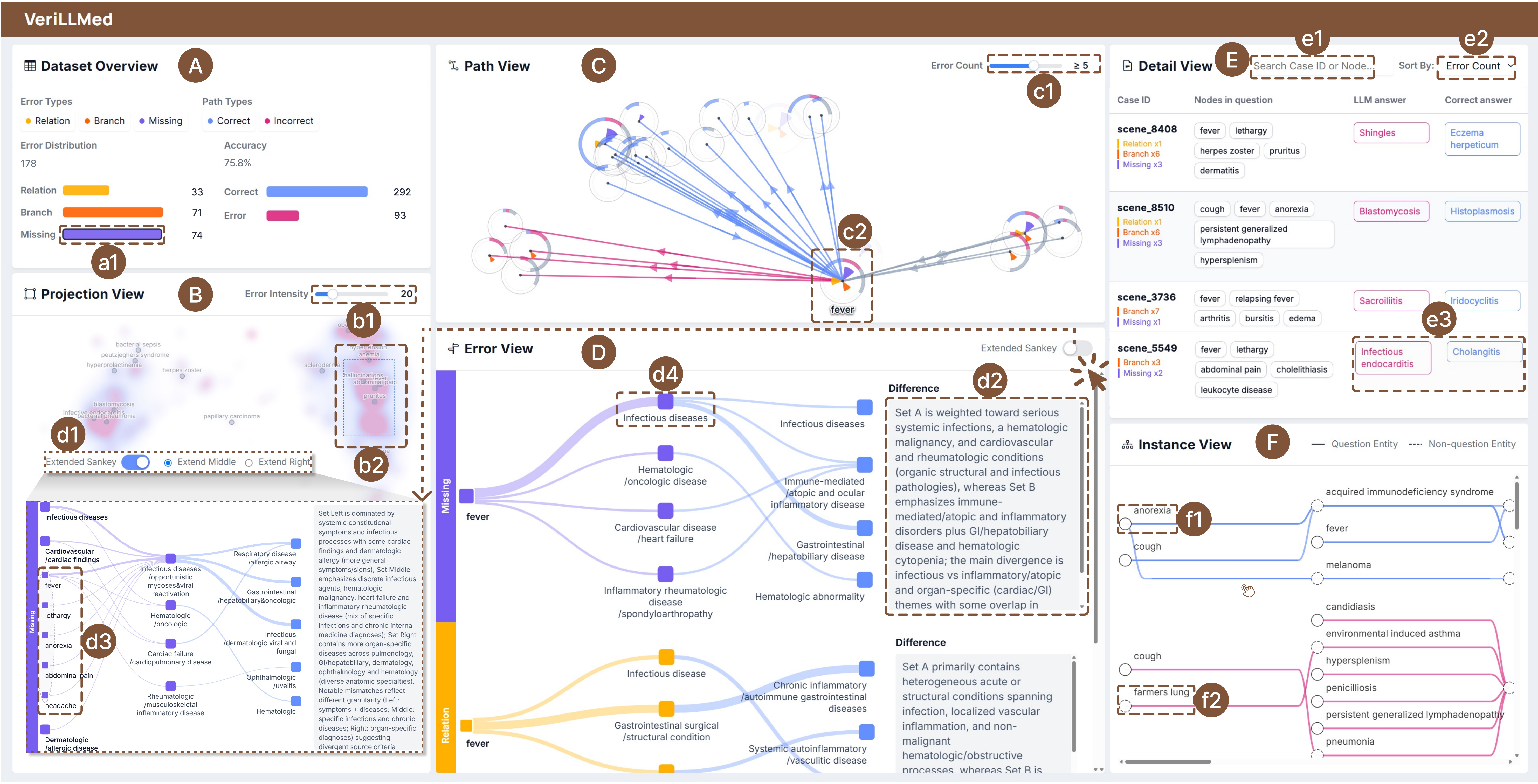}
  \caption{The user interface of \tool. 
    (A) \textit{Dataset Overview} summarizes the distribution of reasoning errors and overall model accuracy. 
    (B) \textit{Projection View} localizes erroneous concept regions in the Bio-KG embedding space. 
    (C) \textit{Path View} compares aggregated local reasoning structures around selected entities. 
    (D) \textit{Error View} summarizes recurrent erroneous reasoning and their corresponding reference reasoning. 
    (E) \textit{Detail View} lists cases associated with the selected node together with case metadata. 
    (F) \textit{Instance View} shows the reference and observed reasoning paths for one selected case.}
  \label{fig:interface}
}

\graphicspath{{figs/}{figures/}{pictures/}{images/}{./}} 

\usepackage{tabu}                      
\usepackage{booktabs}                  
\usepackage{lipsum}                    
\usepackage{mwe}                       

\usepackage{mathptmx}                  

\usepackage[most]{tcolorbox}

\definecolor{errOrange}{HTML}{FFB000}
\definecolor{errCyan}{HTML}{FE6100}
\definecolor{errPurple}{HTML}{785EF0}
\definecolor{correctGreen}{HTML}{648FFF}
\definecolor{observedRed}{HTML}{DC267F}

\newtcbox{\colortag}[2][]{%
  on line,
  nobeforeafter,
  boxrule=0pt,
  colback=#2,
  colframe=#2,
  arc=2.0pt,
  left=6pt,
  right=6pt,
  top=1.4pt,
  bottom=1.4pt,
  boxsep=0pt,
  tcbox raise base=0.1ex,
  valign=center,
  fontupper=\sffamily\bfseries\fontsize{5.0pt}{5.0pt}\selectfont\color{white},
  before upper={\rule[-0.2ex]{0pt}{1.75ex}\raisebox{0.6ex}{\strut}},
}

\begin{document}



\firstsection{Introduction}

\maketitle

Large language models (LLMs) have increasingly demonstrated strong potential in medical diagnosis~\cite{singhal2023multimedqa,singhal2025medpalm2,nori2023gpt4medical, wang2025medkgi, lin2025survey}.
For example, recent studies have shown that medical LLMs can achieve strong performance on challenging benchmark tasks such as USMLE-style medical question answering, and in some settings even approach expert-level performance under structured evaluation protocols~\cite{singhal2023multimedqa,singhal2025medpalm2,nori2023gpt4medical}.
However, given the high-stakes nature of clinical diagnosis, it is critical to inspect LLM behaviors and ensure that their reasoning is reliable and clinically grounded before deployment in real-world clinical practice.
Consequently, developers need to carefully examine potential model errors and debug models to address them.

However, debugging models in medical settings remains challenging.
Although a growing number of benchmarks and assessment frameworks have been proposed to reveal whether LLMs fail on specific medical cases
\cite{jin2021medqa,jin2019pubmedqa,pal2022medmcqa,singhal2025medpalm2,nori2023gpt4medical,lanham2023faithfulness,turpin2023unfaithful},
developers still face a substantial gap in translating these errors into effective improvement strategies.
First,
\textbf{developers often lack sufficient medical domain expertise required to interpret these errors in clinically meaningful terms.}
In practice, developers may collaborate with medical experts to better understand observed incorrect cases.
However, such expert-in-the-loop workflows are costly, difficult to scale, and often impractical for iterative model development.
Second, \textbf{developers struggle to prioritize errors for inspection.}
In practice, models can fail on a large and diverse set of instances, spanning different input types, tasks, and reasoning steps, making it infeasible to inspect each error individually.
Consequently, developers struggle to identify which errors are most critical and which instances deserve the most attention.
Third, even after selecting which errors to inspect, \textbf{developers find it challenging to identify and summarize recurring error patterns across cases.}
Current debugging practices are largely instance-centric, requiring developers to manually examine isolated error cases and their associated reasoning logs. 
Developers find it difficult to connect relevant error cases, which leads to little systematic insight into broader error patterns or the underlying weaknesses of the model.



Prior work has explored algorithmic and interactive approaches for debugging model errors.
Automatic methods 
can expose erroneous behaviors or groups that are hard to notice from aggregate scores alone~\cite{ribeiro2020checklist,yuan2022isea,rajani2022seal}.
However, these identified errors
do not reveal problems in the diagnostic reasoning process.
Developers still struggle to determine why that error is medically problematic and how it should inform concrete refinement actions.
Interactive systems have made important progress on this front.
Tools such as Errudite~\cite{wu2019errudite} and Seq2Seq-Vis~\cite{strobelt2019seq2seqvis}
support error inspection across instances and subgroups, while CommonsenseVIS~\cite{wang2024commonsensevis} enables structured analysis of reasoning patterns with external knowledge support.
However, unlike common prediction tasks, diagnostic reasoning includes multiple steps, and judging whether a reasoning step is plausible requires specialized biomedical knowledge.
Consequently, existing tools do not provide sufficient support for identifying medically meaningful reasoning problems and translating them into insights for model improvement.

These limitations motivate a visual analytics approach integrating external biomedical knowledge for reasoning-level debugging in medical diagnosis.
Rather than treating model reasoning as unstructured text, we map diagnostic reasoning into a structured biomedical concept space based on biomedical knowledge graphs (Bio-KGs). This allows us to analyze reasoning with clinically meaningful entities and relationships, enabling more precise reasoning-level debugging~\cite{su2023ibkh,wu2025medreason}.
Building on this foundation, we present \tool, a visual analytics system for debugging medical LLM diagnostic reasoning through Bio-KGs.
\tool first transforms model outputs into structured reasoning paths and aligns medical entities to a Bio-KG, so that diverse reasoning processes can be compared within a shared graph-based representation.
It then constructs reference paths grounded in the KG and detects three recurring error types that capture distinct forms of diagnostic problems: \textbf{missing errors}, where the model omits key concepts for inferring the correct answer and excluding incorrect options; \textbf{relation errors}, where the model traverses a medically invalid directed relation; and \textbf{branch errors}, where reasoning diverges into concept regions that cannot reach the correct diagnosis.
In summary, this paper makes the following contributions:
\begin{compactitem}
    \item We derive a set of design requirements for reasoning-level debugging in diagnosis-oriented medical QA, clarifying the need for accessible medical grounding, prioritized error triage, recurrent pattern discovery, and case-level validation.
    
    \item We introduce a KG-grounded error taxonomy and corresponding detection pipeline for medical LLM reasoning.
    
    \item We present \tool, a visual analytics system for KG-grounded debugging of medical LLM diagnostic reasoning, which uses biomedical knowledge graphs as an external auditing substrate to support multi-level analysis of reasoning paths, localize recurrent structural errors, and help developers identify actionable targets for model refinement.
    
    \item We demonstrate, through case analyses and expert evaluation, that \tool helps LLM developers identify clinically implausible reasoning and recurring structural error patterns, and derive actionable hypotheses for improving medical LLM reliability.
\end{compactitem}

\section{Related Work}

This section reviews prior work in LLM diagnosis, visual analytics for model debugging, and knowledge graphs in the medical domain.

\subsection{LLM Diagnosis}

A large number of works evaluate LLM outputs through benchmark performance on medical question answering, biomedical reading comprehension, and exam-style clinical reasoning tasks, including MedQA~\cite{jin2021medqa}, PubMedQA~\cite{jin2019pubmedqa}, MedMCQA~\cite{pal2022medmcqa}, and broader medical LLM evaluation methods
\cite{nori2023gpt4medical,singhal2023multimedqa,singhal2025medpalm2,li2024medbench,wang2024cmedbench,chen2025healthcareeval,lacerda2025medicaleval}.
These studies suggest the diagnostic potential of current LLMs and support comparison of models under standardized settings.
However, their analyses are still largely centered on aggregated metrics, such as accuracy, robustness, calibration, or safety-related error rates
\cite{guo2017calibration,ji2023surveyhallucination,bedi2025fidelity, chen2025unmasking}.
Such metrics provide limited support for diagnostic tasks, where model behavior contains a multi-step reasoning process rather than a single prediction.

General LLM research has shown that reasoning process evaluation is necessary for judging whether model-generated rationales are faithful and informative
\cite{lightman2023letsverify,lanham2023faithfulness,turpin2023unfaithful, song2026vizdefender}.
In the medical domain, studies assess clinical reasoning using richer protocols than answer accuracy alone, such as physician evaluation or explicit scoring of written diagnostic reasoning across different stages of clinical decision making
\cite{qiu2025medrbench,mccoy2025clinicalreasoning,zhou2025medthinkbench,tam2024human}.
These studies highlight an important shift from answer-level evaluation toward reasoning process evaluation.

Nevertheless, existing reasoning evaluation methods still fall short of the needs of medical LLM diagnosis.
Most of them evaluate free-form text at the case level, or summarize reasoning quality with statistics, without localizing where the reasoning process structurally fails across many cases\cite{qiu2025medrbench,zhou2025medthinkbench,mccoy2025clinicalreasoning,tam2024human,bedi2025fidelity}.
As a result, they offer limited support for identifying recurrent diagnostic error patterns, comparing erroneous reasoning against medically grounded reasoning, or connecting observed errors to concrete refinement targets.
Our work complements existing reasoning debugging approaches by introducing a KG-based analytic workflow that supports structural comparison of reasoning paths and exposes reasoning errors in clinically meaningful ways.

\subsection{Visual Analytics for LLM Debugging}

Visual analytics has long been used to inspect and debug NLP and LLM models by making predictions, explanations, and error cases interactively explorable\cite{strobelt2018lstmvis,strobelt2019seq2seqvis,wu2019errudite,tenney2020lit,vig2019bertviz,sarti2023inseq,yeh2024attentionviz,tufanov2024lmtt,wang2024vaml}.
Representative systems such as Seq2Seq-Vis~\cite{strobelt2019seq2seqvis}, LIT~\cite{tenney2020lit}, DiLLS~\cite{sheng2026dills}, and BertViz~\cite{vig2019bertviz} allow users to inspect generated outputs, token-level behavior, local explanations, and incorrect transitions.
These systems are valuable for understanding what a model produced and for exploring how outputs vary with prompts or inputs.
However, they mainly organize information around text generation, token interactions, or local model behavior, rather than around a multi-step reasoning process.

Some systems move closer to debugging by supporting error analysis across instances and subgroups.
For example, Errudite~\cite{wu2019errudite} defines reproducible error slices and compares model behavior across data subsets, while CommonsenseVIS~\cite{wang2024commonsensevis} connects model reasoning with external knowledge to support systematic analysis of relational reasoning behavior.
These systems demonstrate the value of analysis based on group-level errors and structured knowledge in the reasoning process\cite{wu2019errudite,wang2024commonsensevis,battogtokh2024semla}.
However, medical diagnosis is a multi-step and domain-intensive process in which the plausibility of one reasoning step often depends on biomedical relations and concept dependencies.
Existing systems provide limited support for identifying recurring reasoning error patterns across cases and clinical contexts.

Our work extends prior work on LLM debugging and proposes a reasoning-level visual analytics workflow for debugging diagnostic errors in a clinically grounded concept space.
We organize model behavior around KG-aligned reasoning paths and explicitly contrast erroneous reasoning with medically grounded paths.
This design allows developers to move from global error summaries to semantically localized concept regions and recurrent error patterns.

\subsection{Knowledge Graphs in the Medical Domain}

Knowledge graphs (KGs) have become an important method for organizing biomedical and clinical knowledge.
They represent diseases, symptoms and other medical concepts as structured medical entities and relations, thereby supporting integration, retrieval, inference, and downstream analytics
\cite{su2023ibkh,cui2025kghealthcare,lu2025biomedicalkgsurvey}.
Such graphs are especially valuable in medicine because diagnostic reasoning is inherently relational, often including associations among diseases, symptoms, findings, and mechanisms.
Accordingly, Bio-KGs have also been incorporated into medical LLM pipelines as external knowledge sources for retrieval, evidence expansion, constrained generation, or reasoning enhancement
\cite{pan2024llmkg_survey,xu2025medklm_survey,yang2025llmkgsurvey,soman2024kgrag,gao2025drknows,wu2025medreason,jia2025medikal}.
For example, MedReason~\cite{wu2025medreason} uses a medical KG to derive factual thinking paths for supervision, while DR.KNOWS~\cite{gao2025drknows} retrieves knowledge paths from UMLS to support diagnosis prediction, and medIKAL~\cite{jia2025medikal} combines candidate disease localization and reranking based on a KG to refine clinical diagnosis.
This line of work shows that structured medical knowledge can improve factuality and support more reliable diagnostic generation.
However, the KG remains an assistive resource for model improvement,
rather than an analytic reference for helping developers understand recurring reasoning errors after generation in a systematic and structural manner.

In contrast, our work leverages the Bio-KG as a post-hoc auditing basis for reasoning analysis.
Rather than using KGs to improve generation, we integrate them into a visual analysis workflow that helps developers interpret diagnostic reasoning through medically grounded structures.
This design is motivated by the fact that many developers are not medical experts and struggle to judge whether an incorrect reasoning chain violates meaningful clinical relationships.
Bio-KGs make such relationships explicit by organizing entities and associations in a structured form.
Thus, our work uses KGs as an analytic reference to help developers audit reasoning traces, identify recurring error patterns, and derive more accessible insights from complex medical errors.

\section{Design Study}
\label{sec:design_study}

Our goal in designing a visual analytics system is to help LLM developers move from answer-level evaluation to reasoning-level debugging in diagnosis-oriented medical QA. In this setting, developers need to assess not only whether a model arrives at the correct diagnosis, but also whether its reasoning is clinically plausible, and whether similar errors recur across semantically related cases. 
To ground the design of our system, we conducted a formative study with five experienced LLM developers (\textbf{\emph{E1--E5}}). \textbf{\emph{E1}} has 6 years of experience and works on LLM-driven healthcare applications. \textbf{\emph{E2}} has 3 years of experience and is a Ph.D. candidate in computational biology and LLMs for science.
\textbf{\emph{E3}} has 1 year of experience and works on LLM-based multi-agent systems for cell inference. \textbf{\emph{E4}} has 3 years of experience and is a product manager and R\&D practitioner at an LLM-for-science company.
\textbf{\emph{E5}} has 1 year of experience and works on medical LLMs 
, involving outpatient reasoning with RAG and multi-agent interaction.
We met with these experts on a biweekly basis for six months to ensure steady progress and to incorporate their feedback throughout the study. This iterative process enabled timely adjustments to our approach.
Our study has received IRB approval. Based on the interviews, we derived five design requirements as follows.


\textbf{R1. Provide accessible external medical knowledge support for reasoning interpretation.}
Based on findings from our formative study, many LLM developers do not have sufficient clinical expertise to determine whether a reasoning step is medically justified. \textbf{\emph{E1}} and \textbf{\emph{E2}} both noted that, in practice, they often need to consult medical references or seek help from domain experts before they can judge whether a model's inference is plausible. However, such expert support is costly and difficult to scale in iterative debugging workflows. Prior work in visual analytics and explainable AI has likewise emphasized the importance of external domain grounding when users analyze complex model behavior in specialized settings \cite{spinner2020explainer,mohseni2021xai,wang2024commonsensevis}. Therefore, the system should provide developers with accessible external medical knowledge support that can serve as a reference for interpreting reasoning, so that they can understand model behavior in clinically meaningful terms even without extensive medical training.

\textbf{R2. Summarize different error types with rates.}
Experts noted that aggregate evaluation is useful for monitoring model behavior, but that debugging requires more informative summaries than overall accuracy alone. In particular, \textbf{\emph{E1}} and \textbf{\emph{E3}} emphasized that model reasoning errors can arise from different kinds of problems. Therefore, separating these error types is important, considering that different error types might call for different debugging strategies. Prior work in visual analytics has similarly shown that concise performance summaries and error typologies are essential for helping users form interpretable hypotheses about model behavior before drilling down into details \cite{spinner2020explainer,strobelt2019seq2seqvis,wang2024commonsensevis}. Therefore, the system should summarize metadata about model performance and error rates, and explicitly distinguish different types of reasoning errors, so that developers can build an initial understanding of how errors are distributed across cases and error categories, and what kinds of problems deserve separate analytical attention.

\textbf{R3. Guide users from aggregate performance metrics to prioritized investigation targets.}
Experts noted that aggregate evaluation is useful for monitoring model behavior, but insufficient for deciding where to investigate next. \textbf{\emph{E1}} and \textbf{\emph{E4}} mentioned that they may observe a performance drop in some portion of the data or that certain types of errors become more frequent, yet still find it difficult to determine which concepts, reasoning structures, or error types should be prioritized for further inspection. As a result, debugging often stalls at the level of noticing that performance is poor, without progressing to more concrete hypotheses about where the underlying problem lies. Therefore, the system should help developers translate high-level performance signals into prioritized investigation targets by connecting global summaries to specific medical fields, entities, and reasoning structures that warrant closer analysis in subsequent debugging steps.

\textbf{R4. Reveal recurrent error patterns across multiple cases.}
The experts stressed that debugging should go beyond isolated erroneous instances and instead support pattern-level analysis across many cases. \textbf{\emph{E4}} and \textbf{\emph{E5}} explained that their current workflow is still largely instance-centric: they may identify a few representative errors, but it remains difficult to determine whether these reflect isolated mistakes or recurring structural patterns shared across many cases. They further noted that comparing erroneous reasoning structures with medically appropriate alternatives across multiple cases is often more useful for guiding refinement than examining single error types in isolation. Prior visual analytics systems likewise show that raw reasoning artifacts are difficult to summarize systematically unless they are transformed into structured and aggregatable representations \cite{strobelt2019seq2seqvis,tenney2020lit,wang2024commonsensevis}. Therefore, the system should help developers identify recurrent error patterns across multiple cases, characterize how erroneous reasoning structures differ from correct ones, and support expansion from one observed error to related families of similar errors.

\textbf{R5. Support detailed inspection and validation of individual cases.}
Although recurring patterns are important, experts also emphasized the need to closely inspect individual cases to validate hypotheses and understand concrete error mechanisms. \textbf{\emph{E4}} noted that, after identifying a suspicious entity or reasoning pattern, developers still need to read representative cases carefully to verify whether the observed pattern is genuinely meaningful and to understand how it manifests in the original question context. This step is currently labor-intensive because developers must manually examine reasoning logs, compare predicted and correct answers, and trace where the reasoning diverges. Prior work in visual analytics similarly highlights the importance of linking aggregate patterns back to instance-level evidence for interpretation and validation \cite{spinner2020explainer,strobelt2019seq2seqvis}. Therefore, the system should support detailed inspection of individual cases, including the involved entities, answers, and reasoning paths, so that developers can validate pattern-level findings and better connect them to concrete debugging actions.

\section{Algorithm}
\label{sec:Algorithm}

\begin{figure*}[t]
    \centering
    \includegraphics[width=0.95\textwidth]{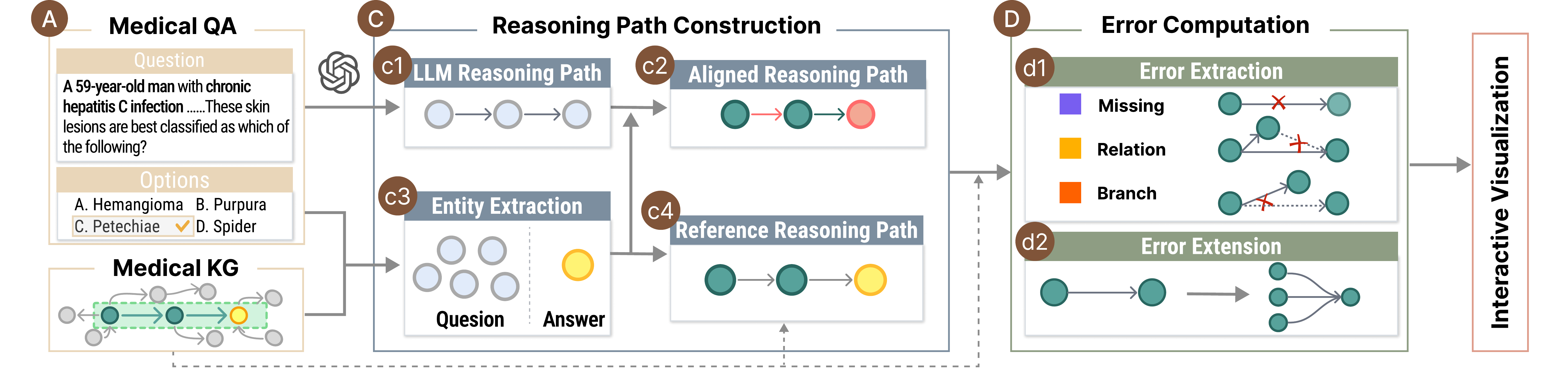}
    \caption{The pipeline of data processing and component extraction. Given a medical QA case, the diagnosis model first predicts the answer together with structured reasoning paths (c1). In parallel, biomedical entities are extracted from the question and the correct answer (c3), and both the model-generated reasoning entities and the QA entities are aligned to the Bio-KG and organized into structural reasoning paths (c2, c4). The two types of paths are then compared based on the Bio-KG to identify \revise{missing, relation, and branch errors}, which are further extended into recurrent error patterns for analysis (d1, d2). The computed reasoning paths and error patterns are finally delivered to the interactive visualization system.}
    \label{fig:pipeline}
\end{figure*}

In this section, we describe the computational pipeline of \tool, including the dataset and model, the structural path construction pipeline, and the error extraction method.

\subsection{System Data and Model}

The analytical foundation of \tool is a collection of diagnosis-oriented medical QA cases,
\(
\mathcal{C}=\{c_1,\dots,c_n\}
\).
Each case
\(
c\in\mathcal{C}
\)
is represented as
\(
c=(q,\mathcal{O},y^{*},\hat{y},P^{g},P^{r})
\),
where
\(
q
\)
is the clinical question,
\(
\mathcal{O}
\)
denotes the candidate diagnosis set,
\(
y^{*}
\)
is the ground-truth diagnosis,
\(
\hat{y}
\)
is the model-predicted diagnosis,
\(
P^{g}
\)
is the set of model-generated reasoning paths, and
\(
P^{r}
\)
is the set of KG-grounded reference reasoning paths.
This representation supports our goal of preserving the inner reasoning structure that leads to the answer. 

\textit{\textbf{Medical QA dataset.}}
Medical QA is a free-form multiple-choice medical QA dataset in which each item typically consists of a clinical question and a set of candidate answers requiring professional biomedical knowledge and multi-step inference
~\cite{jin2021medqa}.
A typical case in such QA datasets is: ``A 27-year-old man presents with abdominal pain and nausea ... Which of the following is the most likely diagnosis?''
with a set of diagnostic options.
We utilize 
MedQA~\cite{jin2021medqa}, a representative medical QA benchmark compiled from professional medical licensing examinations (Fig.~\ref{fig:pipeline}-A).
The original MedQA benchmark covers 61,097 questions~\cite{jin2021medqa}, which 
spans diverse medical content including diseases, symptoms, pathophysiology, laboratory findings, pharmacology, treatment, and clinical management~\cite{jin2021medqa,singhal2023multimedqa}.
In our work, we retain 530 diagnostic-related QAs by filtering MedQA for cases in which the candidate diagnoses and the question are centered on diseases and symptoms, ensuring consistency with the diagnosis-oriented reasoning paths and Bio-KG grounding used in our analysis.

\textit{\textbf{Biomedical knowledge graph.}}
We incorporate a Bio-KG to provide external medical grounding for reasoning analysis (Fig.~\ref{fig:pipeline}-B).
In our system, the knowledge graph is used to align model-generated medical entities to a shared biomedical concept space, to verify whether observed reasoning transitions are supported by biomedical relations.
We choose the integrative Biomedical Knowledge Hub (iBKH)~\cite{su2023ibkh} as the Bio-KG.
iBKH integrates 18 biomedical data sources and contains 2,384,501 entities spanning 11 entity types, together with 45 relation types over 18 kinds of entity pairs~\cite{su2023ibkh}.
Its entity vocabulary covers major biomedical categories such as diseases, symptoms, drugs, genes, pathways, anatomy, molecules, and side effects, while its relations encode diverse biomedical associations across these concept types~\cite{su2023ibkh}.
In our system, we mainly focus on the 19,236 diseases and 1,361 symptoms in iBKH together with medically meaningful relations among them, such as \textit{parent-of}, \textit{resemble}, and \textit{present}, because these entity types constitute the most directly diagnosis-relevant concept space for constructing effective reasoning paths, while other biomedical entity types are less central to the diagnostic task in our work.

\textit{\textbf{Reasoning model.}}
Our system requires a reasoning model that can produce clinically plausible diagnosis predictions with intermediate medical entities and relations for downstream reasoning analysis.
Accordingly, the model should have strong general reasoning ability, reliable instruction following, and stable, structured output behavior. 
Prior medical LLM studies have shown that modern frontier models can already perform strongly on diagnosis-oriented medical QA, making them suitable subjects for reasoning-level debugging~\cite{singhal2023multimedqa,singhal2025medpalm2,nori2023gpt4medical}.
\revise{
We use GPT-5-mini~\cite{openai2025gpt5mini} as the reasoning model in our experiments. GPT-5-mini provides a practical balance of cost, API stability, instruction following, and structured-output controllability, while remaining representative of strong model behavior on diagnosis-oriented medical QA~\cite{singhal2023multimedqa,singhal2025medpalm2,nori2023gpt4medical,openai2026healthcare}. Moreover, for many clinical practitioners and researchers without dedicated GPU resources, API-based models remain the most accessible entry point for deploying such a diagnostic system.}

\subsection{Reasoning Path Construction}

To support reasoning-level analysis, we represent each case as two types of structured reasoning paths (i.e., reasoning paths of the model and reference paths from the KG) in a shared biomedical concept space. 

\subsubsection{Model Reasoning Paths}
\textbf{Reasoning Generation}.
Given a medical QA case
\(
c=(q,\mathcal{O},y^{*},\hat{y},P^{g},P^{r})
\),
the diagnosis model takes the question
\(
q
\)
and answer options
\(
\mathcal{O}
\)
as input, and returns two coupled outputs: the predicted diagnosis
\(
\hat{y}
\)
and the structured reasoning paths
\(
P^{g}
\)
consisting of medical entities and directed relations, which explain the reasoning process of deriving a diagnosis (Fig.~\ref{fig:pipeline}-c1). Formally, we write
\[
(\hat{y},P^{g})=\mathrm{LLM}(q,\mathcal{O}).
\]
\revise{We prompt the LLM to generate structured reasoning paths
because prior work has shown that free-form rationales may be incomplete or unfaithful to the model's actual reasoning process
~\cite{lanham2023faithfulness,turpin2023unfaithful}.}
In parallel, we extract disease and symptom entities from
\(
q
\)
and
\(
\mathcal{O}
\)
through prompt engineering (Fig.~\ref{fig:pipeline}-c3)~\cite{hu2024improving,sivarajkumar2024empirical,agrawal2022fewshot}.
\revise{
We use LLM-based extraction because disease and symptom mentions in medical QAs are often implicit and distributed across the question and answer options. Prior work has shown that LLMs can better exploit contextual cues than dictionary matching alone~\cite{hu2024improving,agrawal2022fewshot}.
}
This allows the original case evidence, the model-generated reasoning paths
\(
P^{g}
\),
and the final diagnosis
\(
\hat{y}
\)
to be represented in a unified biomedical concept space for subsequent path construction and comparison.

\textbf{Entity Alignment with the KG}.\label{sec:KG_alignment}
To make reasoning traces comparable across cases, we align all extracted entities to the Bio-KG (Fig.~\ref{fig:pipeline}-c2, c4). 
This grounding step is applied to both the entities appearing in the model-generated reasoning paths and the entities extracted from the question and answer context.
Let \(e\) denote an extracted entity mention and let \(V_K\) denote the entity set of the KG.
We use a three-stage alignment strategy.
First, we perform exact matching against node names in the KG.
If no exact match is found, we encode \(e\) and KG node names using MedEmbed-large-v0.1, a medical-domain embedding model designed for semantic retrieval and matching in clinical and biomedical text~\cite{balachandran2024medembed}, and compute cosine similarity between the mention and candidate KG nodes.
In the second stage, we select  the most similar candidate whose similarity exceeds an empirical threshold \(\tau=0.9\):
\[
\hat{e}
=
\arg\max_{v\in V_K,\ \cos(f(e),f(v))\ge \tau}
\cos(f(e),f(v)),
\]
where \(f(\cdot)\) denotes the MedEmbed encoder.
We set \(\tau\) as a high-confidence threshold to support more reliable downstream KG-based analysis, based on prior medical KG-grounded reasoning work~\cite{wu2025medreason}, because many biomedical concepts are semantically similar but not equivalent.
\revise{
If no candidate satisfies this threshold, we proceed to a third LLM-based stage alignment, as the LLM can use clinical context to select among semantically similar KG candidates.}
We first retrieve the Top-\(K\) most similar KG candidates according to cosine similarity, and then instruct the LLM to choose the most appropriate concept under the question and answer context:
\[
\hat{e}=\mathrm{LLM}(e,\mathcal{N}_{K}(e),q,\mathcal{O}),
\]
where \(\mathcal{N}_{K}(e)\) denotes the Top-\(K\) candidates of \(e\), and \(q\) and \(\mathcal{O}\) denote the question stem and answer context, respectively.

\subsubsection{Reference Paths}\label{sec:ref_path}
To support reasoning-level comparison, each case additionally requires medically grounded reference paths.
We construct these paths with a KG-based procedure adapted from recent medical reasoning generation work~\cite{wu2025medreason}, but use the resulting graph paths directly as structural references for structural comparison and reasoning error analysis (Fig.~\ref{fig:pipeline}-c4).

For a case \(c\), let \(X_c\) denote the set of disease and symptom entities extracted from the question and answer context, and let \(y_c^{*}\) denote the correct diagnosis.
After aligning \(X_c\) and \(y_c^{*}\) with the Bio-KG, we search the graph for the shortest directed paths that connect entities from the question to the correct answer entity.
\revise{
We then use the LLM to prune the candidate shortest paths based on the original question~\cite{wu2025medreason}, removing paths that are valid in the KG structure but clinically irrelevant to the specific case semantics.
}
For each grounded question entity \(x\in X_c\), let
\[
\Pi(x,y_c^{*})=\{\pi_1,\ldots,\pi_m\}
\]
denote the set of shortest KG paths from \(x\) to \(y_c^{*}\).
The retained paths are
\[
\widetilde{\Pi}(x,y_c^{*})=\mathrm{LLM}\!\left(\Pi(x,y_c^{*}),q,\mathcal{O}\right).
\]
The final reference paths are formed by aggregating the retained paths across all relevant question entities:
\[
P^{r}=\bigcup_{x\in X_c}\widetilde{\Pi}(x,y_c^{*}).
\]

This procedure yields a compact set of KG-grounded biomedical transitions that serves as the reference reasoning structure for case \(c\).

\subsection{Error Definition and Pattern Extraction}

We define reasoning errors by comparing the observed model reasoning paths
\(
\tilde{P}^{g}
\)
with the reference paths
\(
P^{r}
\)
in the same biomedical concept space.
Here, \(\tilde{P}^{g}\) denotes the aligned model reasoning paths derived from \(P^{g}\) in Section~\ref{sec:KG_alignment}.
Our goal is 
to decompose structural deviations into a set of interpretable error types and retrieve more related errors that can support targeted debugging.

\subsubsection{Error Definition}\label{sec:error_def}

Let
\(
G_{K}=(V_{K},E_{K})
\)
denote the Bio-KG.
To evaluate whether one concept can support another concept in the reasoning process, we use path reachability.
Specifically, we define a directed graph
\(
G_{K}^{\pm}
\)
from the Bio-KG in which biomedical relations such as \textit{parent-of} keep their original direction, while relations based on similarity such as \textit{present} and \textit{resemble} are treated as bidirectional.
For two KG nodes \(u,v\in V_{K}\), we write
\(
\mathrm{Reach}(u,v)=1
\)
if there exists a directed path from \(u\) to \(v\) in
\(
G_{K}^{\pm}
\),
and
\(
\mathrm{Reach}(u,v)=0
\)
otherwise.

Considering a single reasoning step from entity \(e_A\) to entity \(e_B\) (Fig.~\ref{fig:pipeline}-d1), 
we distinguish three types of structural errors, arising from the starting node, the connecting edge, and the ending node:

\begin{compactitem}
    \item 
\textbf{Missing error.} If \(e_A\) does not appear in the observed paths but is a key entity in the reference reasoning, then the model has missed a diagnostically useful entity.
A missing error captures the omission of a key concept that appears in the reference reasoning but not in the observed reasoning.
Such a concept should support the correct diagnosis while not supporting the model's incorrect prediction.
Let
\(
V(P^{r})
\)
and
\(
V(\tilde{P}^{g})
\)
denote the node sets of the reference and observed paths, respectively.
For a reference node
\(
M \in V(P^{r}) \setminus V(\tilde{P}^{g})
\),
we define a missing error whenever
\(
\mathrm{Reach}(M,y^{*})=1
\)
and
\(
\mathrm{Reach}(M,\hat{y})=0
\).

\item
\textbf{Relation error.} 
The model connects entity \(e_A\) to an entity \(e_B\) that is not reachable from \(e_A\) under biomedical knowledge, indicating an invalid transition. A relation error occurs when the model advances from one entity to another, but the transition is not supported by the KG under the reachability criterion above.
For an observed step
\(
(e_A,e_B)\in \tilde{P}^{g}
\),
we define a relation error whenever
\(
\mathrm{Reach}(e_A,e_B)=0
\).

\item
\textbf{Branch error.} 
\(e_A\) connects to \(e_B\) through a valid Bio-KG path, and \(e_A\) can reach the correct diagnosis but \(e_B\) cannot under biomedical knowledge.
A branch error occurs when the model follows an entity that is reachable under current reasoning context, yet that concept does not support the correct diagnosis.
Let
\[
\mathrm{Anc}(y^{*})=\{v\in V_{K}\mid \mathrm{Reach}(v,y^{*})=1\}
\]
denote the set of KG nodes from which the correct diagnosis \(y^{*}\) is reachable.
For an observed grounded step
\(
(e_A,e_B)\in \tilde{P}^{g}
\),
we define a branch error whenever
\(
e_A\in \mathrm{Anc}(y^{*}) \land e_B\notin \mathrm{Anc}(y^{*})
\).

\end{compactitem}

For each case \(c\), after grounding the observed and reference reasoning paths to the KG, we compute three error sets:
\[
\mathcal{E}_{c}^{\mathrm{rel}}
=
\{(e_A,e_B)\in \tilde{P}^{g}\mid \mathrm{Reach}(e_A,e_B)=0\},
\]
\[
\mathcal{E}_{c}^{\mathrm{br}}
=
\{(e_A,e_B)\in \tilde{P}^{g}\mid \mathrm{Reach}(e_A,y^{*})=1 \land \mathrm{Reach}(e_B,y^{*})=0\},
\]
and
\[
\mathcal{E}_{c}^{\mathrm{miss}}
=
\{M\in V(P^{r})\setminus V(\tilde{P}^{g})\mid \mathrm{Reach}(M,y^{*})=1,\ \mathrm{Reach}(M,\hat{y})=0\}.
\]
The corresponding error intensities are counted as
\[
N_{c}^{\mathrm{rel}}=|\mathcal{E}_{c}^{\mathrm{rel}}|,\qquad
N_{c}^{\mathrm{br}}=|\mathcal{E}_{c}^{\mathrm{br}}|,\qquad
N_{c}^{\mathrm{miss}}=|\mathcal{E}_{c}^{\mathrm{miss}}|.
\]
These counts provide a compact summary of how the model's reasoning fails on a given case and serve as the basis for later ranking and retrieval.

\subsubsection{Error Pattern Extraction}\label{sec:err_pattern}
To support pattern-level analysis beyond a single erroneous transition, 
\revise{
we define a pattern as a group of related errors that share 
either the same erroneous target nodes or the same reference target nodes (Fig.~\ref{fig:pipeline}-d2).
}
\revise{
For each error type \(\tau \in \{\mathrm{Missing}, \mathrm{Relation}, \mathrm{Branch}\}\), we denote the erroneous node pairs and their corresponding reference node pairs as \(\mathcal{E}^{\tau}_{\mathrm{err}}\) and \(\mathcal{E}^{\tau}_{\mathrm{ref}}\), respectively.
We further define a correspondence set
\[
\mathcal{M}^{\tau}=\{((s,t_{\mathrm{err}}),(s,t_{\mathrm{ref}})) \mid (s,t_{\mathrm{err}})\in \mathcal{E}^{\tau}_{\mathrm{err}}, (s,t_{\mathrm{ref}})\in \mathcal{E}^{\tau}_{\mathrm{ref}}\},
\]
where each element links an erroneous reasoning pair with its reference counterpart from the same source node.
Given an initial correspondence \(m_0=((s_0,t^0_{\mathrm{err}}),(s_0,t^0_{\mathrm{ref}}))\), we support two expansion operations. The first retrieves correspondences that share the erroneous target:
\[
\mathcal{P}^{\tau}_{\mathrm{err}}(m_0)=
\{m\in \mathcal{M}^{\tau} \mid
\pi_{\mathrm{err}}(m)=t^0_{\mathrm{err}}\},
\]
and the second retrieves correspondences that share the reference target:
\[
\mathcal{P}^{\tau}_{\mathrm{ref}}(m_0)=
\{m\in \mathcal{M}^{\tau} \mid
\pi_{\mathrm{ref}}(m)=t^0_{\mathrm{ref}}\}.
\]
where \(\pi_{\mathrm{err}}\) and \(\pi_{\mathrm{ref}}\) return the target nodes of the erroneous and reference pairs in a correspondence, respectively.
The system includes either the erroneous or reference pairs in \(\mathcal{P}^{\tau}(m_0)\), enabling users to inspect how an initial error connects to a broader family of related reasoning patterns.
}
To convert such expanded structures into more concise, human-readable semantic categories or comparative explanations,
we further instruct the LLM to summarize the semantic difference between the erroneous node set and the reference node set.
Given the two node sets
\(
\mathcal{B}(e_A)
\)
and
\(
\mathcal{D}(e_A)
\),
the LLM first assigns high-level semantic categories to the nodes in each set and then produces a comparative summary of their major distinctions:
\[
(\mathcal{C}_{\mathrm{err}},\mathcal{C}_{\mathrm{ref}},S)
=
\mathrm{LLM}(\mathcal{B}(e_A),\mathcal{D}(e_A),q,\mathcal{O}),
\]
where
\(
\mathcal{C}_{\mathrm{err}}
\)
and
\(
\mathcal{C}_{\mathrm{ref}}
\)
denote the high-level categories for the erroneous and reference node sets, respectively, and
\(
S
\)
denotes the generated summary of their major semantic differences.

\section{User Interface}

\begin{figure}[t]
    \centering
    \includegraphics[width=\linewidth]{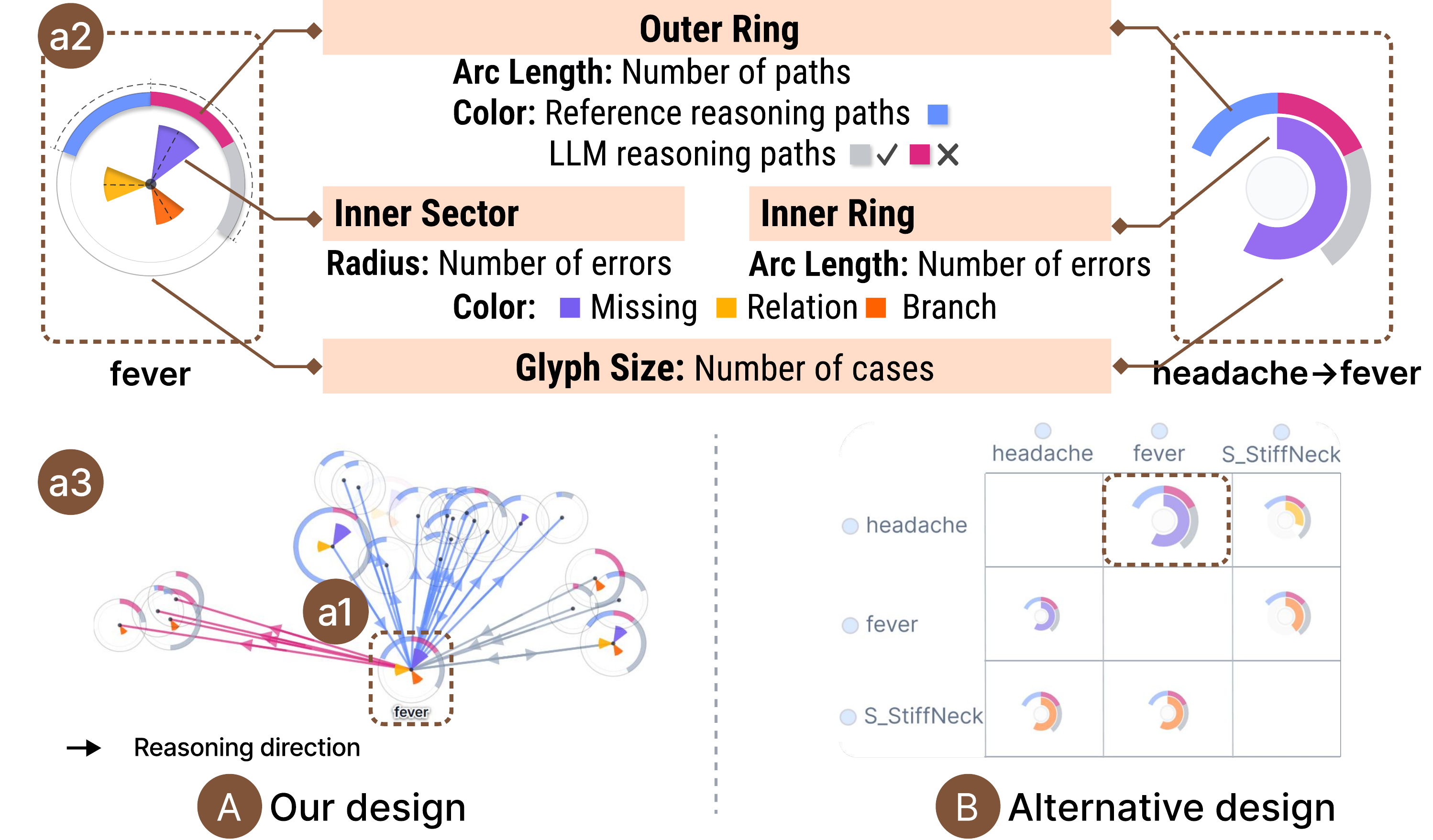}
    \caption{(A) The glyph design of the \textit{Path View}, based on the Bio-KG embedding space. Node positions indicate their connections in the KG. (B) We also considered a matrix-based alternative, where each cell represents a transition from a row entity to a column entity and contains a similar glyph summarizing path participation and error composition.}
    \label{fig:alternative_path}
\end{figure}


\tool provides a multi-level interface for debugging medical diagnostic reasoning with Bio-KG support.
Users begin with the \textit{Dataset Overview} (Fig.~\ref{fig:interface}-A) to identify which error type deserves attention, and then use the \textit{Projection View} (Fig.~\ref{fig:interface}-B) to localize problematic biomedical concept regions.
From there, the \textit{Path View} (Fig.~\ref{fig:interface}-C) helps inspect aggregated reasoning structures around selected entities, while the \textit{Error View} (Fig.~\ref{fig:interface}-D) summarizes recurrent error patterns and their corresponding correct alternatives.
Users can examine related cases in the \textit{Detail View} (Fig.~\ref{fig:interface}-E) and inspect complete reference and observed reasoning paths for one case in the \textit{Instance View} (Fig.~\ref{fig:interface}-F).
\revise{
Throughout the system,
\cl{R2.2, R3.1}
missing errors, relation errors, and branch errors are encoded in purple \colortag{errPurple}{missing}, yellow \colortag{errOrange}{relation}, and orange \colortag{errCyan}{branch}, respectively.
Reference reasoning paths and incorrect reasoning from the model are encoded in blue \colortag{correctGreen}{correct} and magenta \colortag{observedRed}{incorrect}.
}

\subsection{Dataset Overview}

The \textit{Dataset Overview} (Fig.~\ref{fig:interface}-A) helps users quickly assess overall model behavior and choose which type of reasoning error to investigate first.
The left side summarizes the total number of error cases and the distribution of the three error types described in~\cref{sec:error_def}.
The right side reports the overall accuracy along with the number of correct and incorrect cases (\textbf{R2}, \textbf{R3}).
\cll{R2.5} 
Users can select one error type (Fig.~\ref{fig:interface}-a1) to filter the subsequent views so that the \textit{Projection View} and \textit{Path View} focus only on entities and cases associated with the chosen error type.
This interaction helps users derive different debugging implications from different error types.

\subsection{Projection View}

The \textit{Projection View} (Fig.~\ref{fig:interface}-B) helps users move from global error statistics to semantically localized investigation targets in the biomedical concept space.
To construct this view, we first compute node embeddings over the Bio-KG using node2vec~\cite{grover2016node2vec}, which learns representations that preserve graph neighborhood structure.
We then project these embeddings into two dimensions with t-SNE~\cite{vandermaaten2008tsne}. Thus nearby nodes correspond to related biomedical regions (\textbf{R1}).
The view contains two layers.
The first layer is a heatmap that visualizes error intensity, 
which is the accumulation of selected errors in the biomedical concept space.
The magenta background areas indicate regions with high concentrations of selected errors (\textbf{R3}).
The second layer shows top-\(K\) nodes with the highest error counts under the current filter.
A slider above the view allows users to control this threshold and focus on more error-prone entities (Fig.~\ref{fig:interface}-b1).
Users can brush over one hotspot to define a local region of interest.
The brushed entities are then presented in the \textit{Path View} for structural reasoning inspection (Fig.~\ref{fig:interface}-b2).

\subsection{Path View}
\label{sec:pathview}

After selecting a group of entities from the medical concept region in \textit{Projection View} (Fig.~\ref{fig:interface}-C), users can compare aggregated reasoning structures around selected entities and identify which nodes warrant deeper and more targeted pattern analysis.
This panel visualizes the reasoning neighborhood of selected entities.
Each node is rendered as a circular glyph, with links summarizing how the selected entities are connected to subsequent entities in the reference and observed reasoning paths (Fig.~\ref{fig:alternative_path}-A).
To preserve users' Bio-KG context when moving from regional selection to detailed inspection, we place the selected entities using the same projection layout as in the \textit{Projection View} (\textbf{R1}).
The view also supports zooming and panning, allowing users to further inspect error-intensive regions or entities.

Each node is shown as a circular glyph with two layers (Fig.~\ref{fig:alternative_path}-a2).
The outer layer summarizes path participation.
\cl{R3.4} 
\revise{It first splits the circle along a fixed vertical axis: the left half shown in blue represents participation in reference reasoning paths, while the right half encodes its participation in observed model reasoning paths. The right half is further divided into magenta and gray arcs to distinguish erroneous and non-error model reasoning, respectively.}
For each arc, larger arc spans indicate higher frequencies.
The inner sectors encode error information: the radius of each colored sector indicates the intensity of the corresponding error type for that node.
The size of the glyph encodes the frequency of the entity's appearance in all cases.
Users can click a node to highlight its directly connected neighborhood with links between entities (Fig.~\ref{fig:alternative_path}-a3).
Similarly, blue paths denote correct reasoning steps for reference, while magenta and gray paths denote incorrect and correct reasoning steps from the model, respectively.
Arrowheads on the links indicate reasoning direction.
This interaction reduces visual clutter and turns the view into a focused inspection tool for one entity at a time.
Through this design, the glyph conveys both reasoning-level and error-level information (\textbf{R3}, \textbf{R4}).
Once a node is selected, its incoming and outgoing connections are emphasized, allowing users to inspect whether the reasoning step is dominated by one error type, whether the node mainly appears in incorrect reasoning, and which downstream concepts recur most often.
A slider above the view controls the minimum error intensity for displayed nodes (Fig.~\ref{fig:interface}-c1) for analysis on the most problematic entities.

\textbf{Alternative design.}
We also considered a design based on a matrix (Fig.~\ref{fig:alternative_path}-B), in which each cell represents the reasoning step from the row node to the column node.
The outer layer of the glyph shows the participation in the reference and observed reasoning paths, while the inner layer encodes the proportion of the three error types.
However, using a matrix is not space-efficient especially in diagnostic reasoning, which involves many entities.
More importantly, it makes it difficult for users to directly compare the downstream correct and incorrect entities in the reasoning process and weakens users’ perception of the medical relationships among entities in the KG.

\subsection{Error View}
\label{sec:errorview}

\begin{figure}[t]
    \centering
    \includegraphics[width=\linewidth]{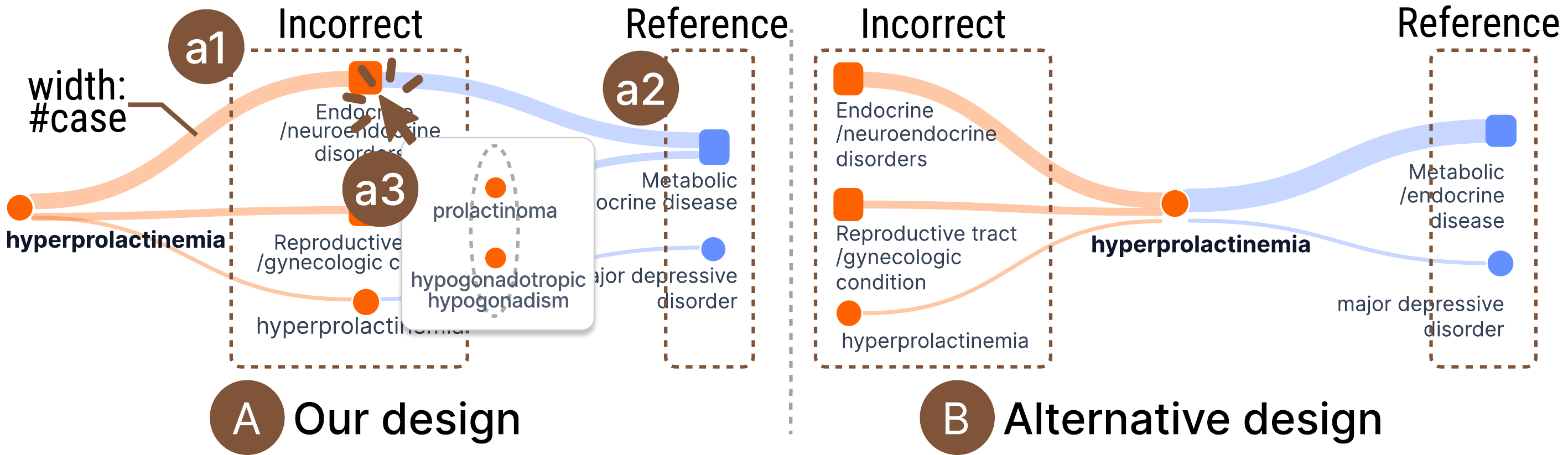}
    \caption{(A) Our design uses a comparative Sankey layout to summarize recurrent error patterns together with their corresponding reference paths under high-level semantic grouping. (B) We also considered a symmetric Sankey alternative, which places the selected node in the center and arranges the erroneous and reference reasoning on the two sides.}
    \label{fig:alternative_error}
\end{figure}

The \textit{Error View} (Fig.~\ref{fig:interface}-D) summarizes recurrent error patterns around a selected node and compares erroneous reasoning against corresponding reference paths.
Once a node is selected in the \textit{Path View}, the \textit{Error View} displays erroneous reasoning associated with that node in a Sankey layout (Fig.~\ref{fig:alternative_error}-A).
The purpose of this view is to move beyond isolated reasoning paths and reveal similar error pattern families that recur across cases.
For missing errors, since the selected entity is omitted from the model reasoning, the view compares the incorrect answers not ruled out by that entity and the correct answers that should have been supported by it 
(Fig.~\ref{fig:interface}-D) (\textbf{R1}).
For relation and branch errors, the left side represents the selected anchor node, the middle column summarizes downstream nodes in the error set from LLM reasoning (Fig.~\ref{fig:alternative_error}-a1), and the right column summarizes the corresponding downstream nodes in the correct set from reference reasoning introduced in~\cref{sec:err_pattern} (Fig.~\ref{fig:alternative_error}-a2).
The widths of the flows represent how many cases contain the corresponding reasoning step.
Wider links indicate more recurrent patterns and deserve higher debugging priority.

The interface also supports error pattern expansion described in~\cref{sec:err_pattern}.
Starting from an anchor entity, users can expand based on either the error set 
\cll{R2.6} 
\revise{in the middle} or the correct set \revise{on the right} (Fig.~\ref{fig:interface}-d1).
If an error of current entity is identified, users can retrieve additional errors whose subsequent reasoning entity belongs to the error set or the correct set, thereby revealing a larger family of related erroneous transitions (\textbf{R4}).
This interaction allows users to follow not just one erroneous step, but a group of related transitions and their reference alternatives.
Additionally, to help users make a more efficient comparison, 
we group the entities in the correct and error set into high-level semantic groups summarized by the LLM in~\cref{sec:err_pattern}.
Users can recognize pattern-level shifts easily without medical domain knowledge.
To support detailed verification, users can then click to expand a category into its detailed entities (Fig.~\ref{fig:alternative_error}-a3), revealing the concrete medical entities underlying the summarized pattern.
The panel also provides a difference summary between the erroneous and the reference reasoning (Fig.~\ref{fig:interface}-d2).
It explains the core distinction between the two sets in concise language.
Users can interpret why the two sides differ medically and gain deeper insight into possible model refinements.

\textbf{Alternative design.}
We also considered an alternative using a similar Sankey layout (Fig.~\ref{fig:alternative_error}-B), in which the selected node is placed in the center, while the error set and the reference set are positioned on the left and right sides, respectively.
However, this design weakens the correspondence between erroneous and correct reasoning.
Users have difficulty determining which erroneous transitions should be compared against which reference transitions.
As a result, it becomes harder 
to derive insight through direct comparison between the two reasoning sets.
We therefore adopted the current comparative layout.

\subsection{Detail View}

The \textit{Detail View} (Fig.~\ref{fig:interface}-E) helps users verify whether a suspicious node or error pattern is supported by concrete case evidence.
This panel lists all cases associated with the node selected in the \textit{Path View}.
For each case, it shows the metadata, including the entities in the medical question, the counts and types of errors involved, the model prediction, and the correct answer (\textbf{R5}).
This allows users to quickly assess the background information of related cases.
Users can also click a case row to open the original medical QA in a pop-up window.
The cases can be searched by related entities (Fig.~\ref{fig:interface}-e1) or sorted by error count (Fig.~\ref{fig:interface}-e2) so that users can inspect the most severe or representative errors for targeted validation.
Users can click one case to inspect its complete reasoning paths in the \textit{Instance View}.

\subsection{Instance View}

The \textit{Instance View} (Fig.~\ref{fig:interface}-F) helps users inspect, at the case level, exactly how the model-generated reasoning differs from the KG-grounded reference reasoning.
After one case is selected from the \textit{Detail View}, this panel visualizes its complete reference and observed reasoning paths together with the relations between entities (\textbf{R5}).
To help users distinguish whether an error stems from misusing available evidence or from introducing problematic intermediate entities, 
the view uses solid outlines (Fig.~\ref{fig:interface}-f1) and dashed outlines (Fig.~\ref{fig:interface}-f2) to represent entities that are directly mentioned or not mentioned in the original question.
This view serves as the final validation step in the workflow: users can check whether the pattern identified in the aggregated views is genuinely reflected in the case-level reasoning process.


\section{Case Study}
\label{sec:casestudy}

\begin{figure}[t]
    \centering
    \includegraphics[width=\linewidth]{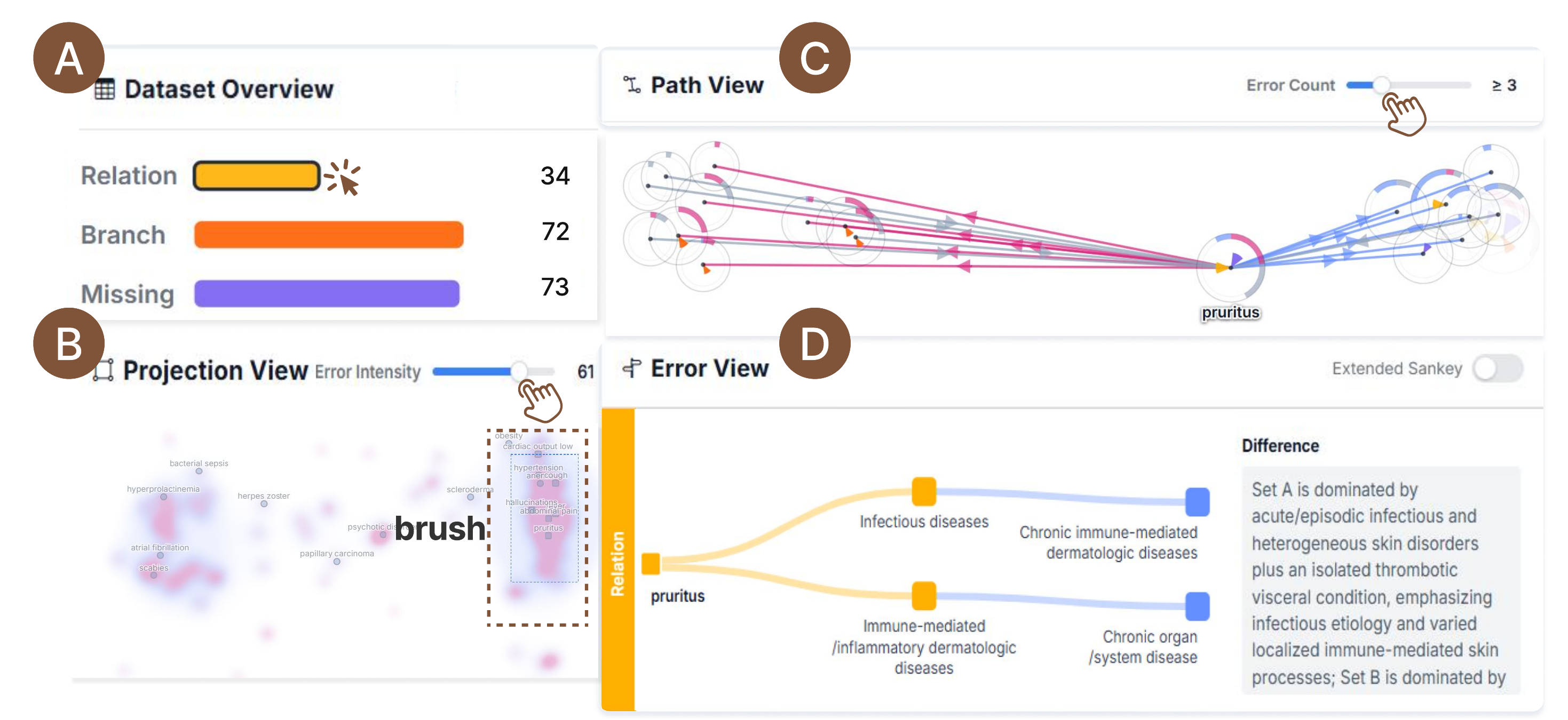}
    \caption{Case 2: Relation bias around \textit{dermatologic symptoms}.
    (A) \textbf{\emph{E2}} selected relation errors to focus on clinically inappropriate relational transitions.
    (B) The expert brushed a relation-error-intensive region for further inspection.
    (C) \textit{Pruritus} appeared as the most erroneous entity, whose highlighted paths showed observed reasoning repeatedly shifting toward distant \textit{dermatologic} associations rather than nearby \textit{systemic} conditions.
    (D) The \textit{Error View} supported this path-level finding by contrasting erroneous \textit{infectious/skin-related} concepts with reference \textit{systemic autoimmune} and \textit{chronic non-dermatologic} concepts.
    }
    \label{fig:case2}
\end{figure}

We present two representative analytic walkthroughs to illustrate how \tool supports real debugging workflows and leads to refinement hypotheses. These walkthroughs are not controlled effectiveness studies.
We invited \textbf{\emph{E1}} and \textbf{\emph{E2}}, who had continuously participated in the formative design and development of \tool\ (\cref{sec:design_study}), to use the system and reflect on its utility through end-to-end case analysis.
Before the session, we introduced the analytical workflow of \tool.
They then used the system to analyze model behavior on the MedQA~\cite{jin2021medqa} benchmark in a think-aloud setting, while we recorded their observations and follow-up explanations.
After the session, we reviewed the discovered patterns with them, confirmed the resulting interpretation, and further examined the derived hypothesis through prompt engineering outside the system.
We summarized two representative cases that capture how the system supported reasoning-level debugging and the formulation of refinement hypotheses.
{This was covered by the IRB approval.}

\subsection{Case 1: Missing Clinically Important but Nonspecific Symptoms}
\label{sec:case1}

\textbf{Overview and Regional Summary (R1, R2).}
\textbf{\emph{E1}} first inspected the overall accuracy and the three error types, and selected the \textit{missing error} type in the \textit{Dataset Overview} (Fig.~\ref{fig:interface}-a1), to examine whether the model fails to use clinically informative evidence in the question.
Subsequent views were updated to focus on missing diagnostic concepts.
He then turned to the \textit{Projection View} (Fig.~\ref{fig:interface}-B), where several error-intensive hotspots were presented.
Among them, he brushed the prominent upper-right region to examine the entities and errors in this medical field (Fig.~\ref{fig:interface}-b2).
After brushing, the \textit{Dataset Overview} statistics for the selected region showed that \textit{missing errors} were the dominant error type, and that the model accuracy within this region was only \textbf{75.8\%}, lower than the \textbf{81.5\%} global accuracy.
This suggested to \textbf{\emph{E1}} that the selected region contained a recurrent omission pattern rather than a few isolated case-specific errors.

{\textbf{Path-level Inspection of a Suspicious Anchor Node (R2, R3)}}.
To narrow the analysis to the most problematic entities in this region, \textbf{\emph{E1}}
kept entities with at least five \textit{missing errors} in the \textit{Path View} (Fig.~\ref{fig:interface}-c1), to further analyze representative omitted entities.
Among them, the symptom node \textit{fever} stood out as a representative missed entity, as suggested by its size and inner error sectors (Fig.~\ref{fig:interface}-c2).
This node was analytically meaningful because \textit{fever} is common and diagnostically important, yet by itself often remains nonspecific.
The local path structure around \textit{fever} also connected to multiple downstream concepts, making it a suitable anchor for recurrent pattern analysis.

\textbf{Recurrent Missing-error Pattern Analysis (R3, R4).}
After selecting \textit{fever}, \textbf{\emph{E1}} examined the \textit{Error View} (Fig.~\ref{fig:interface}-D).
The Sankey comparison and difference summary of \textit{missing errors} showed a clear inconsistency between the erroneous and reference reasoning.
Through the high-level categories, \textbf{\emph{E1}} noticed the downstream incorrect answers associated with \textit{missing errors} were dominated by \textit{infectious diseases} (Fig.~\ref{fig:interface}-d4).
On the reference reasoning side, however, the corresponding correct entities were more often grouped into \textit{immune-mediated} or \textit{atopic inflammatory diseases}, together with more \textit{organ-specific themes}.
This pattern suggested to \textbf{\emph{E1}} that
the model might fail to combine \textit{fever} with other clinical findings to rule out \textit{infectious etiologies}, while insufficiently considering reasoning paths from \textit{fever} toward \textit{noninfectious} but clinically appropriate diagnoses.
When he further expanded the pattern based on \textit{fever} (Fig.~\ref{fig:interface}-d1), he found that the node categories in the \textit{Error View} further unfolded into symptoms such as \textit{abdominal pain} and \textit{headache}, revealing a broader family of \textit{nonspecific systemic symptoms} that showed similar behavior (Fig.~\ref{fig:interface}-d3).
This expansion indicated that the issue was recurrent across a broader family of \textit{nonspecific systemic symptoms} rather than unique to a single node or a small number of isolated cases.

\textbf{Case-level Verification in the Detail and Instance Views (R5).}
To validate whether the aggregated pattern reflected actual case reasoning, \textbf{\emph{E1}} next inspected the related cases in the \textit{Detail View} (Fig.~\ref{fig:interface}-E).
He observed that many of these cases contained question entities such as \textit{fever} and \textit{abdominal pain}, but the model predictions tended toward \textit{infectious diagnoses} even when the correct answers were not (Fig.~\ref{fig:interface}-e3).
He then inspected representative cases in the \textit{Instance View} (Fig.~\ref{fig:interface}-F) and compared the LLM reasoning paths with the reference paths.
Across these cases, the reference reasoning usually started from \textit{nonspecific systemic symptoms}, and combined them with additional \textit{immune-related} or \textit{organ-specific} findings to reach the correct diagnosis.
By contrast, the LLM reasoning often failed to integrate these entities with the corresponding pathways, and instead moved toward more generic infectious interpretations.
These case-level observations were consistent with the recurrent pattern summarized in the \textit{Error View} and further strengthened \textbf{\emph{E1}}'s refinement hypothesis.

\textbf{From Visual Insight to Prompt Intervention (R3, R4).}
Based on this finding, \textbf{\emph{E1}} designed a small prompt-based follow-up study to examine whether the discovered pattern could translate to a plausible refinement hypothesis.
Outside the system, he constructed few-shot examples from MedQA~\cite{jin2021medqa} and MedMCQA~\cite{pal2022medmcqa} that shared the same high-level structure: the question included \textit{nonspecific systemic symptoms}, the options contained \textit{infectious} or \textit{inflammatory} candidates, and the correct answer was a \textit{noninfectious} disease.
These examples were selected from data outside the analyzed cases, which are provided in the supplementary material.
He then ran the model again with these examples added to the prompt.
Under this limited prompt modification, the eight previously incorrect cases associated with this error pattern were corrected, and the benchmark accuracy on the analyzed cases increased from \textbf{81.5\%} to \textbf{84.3\%}.

\subsection{Case 2: Relation Bias Around Dermatologic Symptoms}
\label{sec:case2}

\textbf{Overview and Regional Summary (R1, R2).}
Another expert \textbf{\emph{E2}} was interested in \textit{relation errors} because they indicate that the model connects medical entities through clinically inappropriate relations even when the entities themselves appear relevant semantically.
He therefore selected the \textit{relation error} type in the \textit{Dataset Overview} (Fig.~\ref{fig:case2}-A), which updated the coordinated views to focus on incorrect relational transitions.
He then examined the \textit{Projection View} (Fig.~\ref{fig:case2}-B) and brushed the prominent error-intensive region on the right side.

\textbf{Path-level Evidence of a Localized Reasoning Bias (R2, R3).}
To further narrow the analysis, \textbf{\emph{E2}} retained only the nodes with the highest numbers of \textit{relation errors} in the \textit{Path View} (Fig.~\ref{fig:case2}-C).
Among them, \textit{pruritus} stood out as the most problematic node, with the largest erroneous reasoning count indicated by its outer layer.
After selecting this node, he compared the highlighted observed and reference paths around it.
He observed that the reference reasoning paths, shown in blue, remained largely concentrated within the medical neighborhood of \textit{pruritus}, whereas many observed reasoning paths extended toward a more distant left-side region.
\textbf{\emph{E2}} noticed that rather than diffusing broadly, the model reasoning repeatedly moved from \textit{pruritus} toward \textit{dermatologic associations} although they are distant in medical space, while the reference reasoning advanced to nearby \textit{systemic autoimmune} and \textit{chronic organ-related} conditions.
This pattern suggested to \textbf{\emph{E2}} that although \textit{pruritus} is a \textit{dermatologic symptom}, in diagnostic reasoning it can also function as an entry point to broader \textit{systemic autoimmune} or \textit{chronic organ-related} conditions.
The model might overemphasize the association of \textit{pruritus} with semantically similar entities such as \textit{dermatologic diseases}, while underestimating its association with \textit{systemic diseases} that were more clinically consistent with the medical knowledge and reference reasoning.

\textbf{Pattern verification in the Error View (R3, R4).}
\textbf{\emph{E2}} then examined the \textit{Error View} (Fig.~\ref{fig:case2}-D) to interpret this path-level pattern at a higher semantic and category-based level.
The Sankey comparison for \textit{relation errors} showed that the downstream error side was dominated by \textit{acute or episodic infectious} and \textit{heterogeneous skin disorders}, whereas the reference side was dominated by more \textit{chronic systemic inflammatory or autoimmune conditions}, as well as \textit{chronic non-dermatologic organ disease}.
This summary corresponds to the finding in the \textit{Path View}: the model appeared to treat \textit{pruritus} as a cue for localized \textit{dermatologic} reasoning, while the reference reasoning more often linked it to \textit{systemic disease} processes and \textit{chronic conditions}.

\textbf{From Visual Insight to Prompt Intervention (R3, R4).}
Similarly, based on this finding, \textbf{\emph{E2}} selected a small set of few-shot examples outside the analyzed cases whose questions contained \textit{pruritus} or similar \textit{dermatologic symptoms}, while the correct answers belonged to \textit{systemic autoimmune diseases} or \textit{chronic non-cutaneous organ diseases}. The selected examples are provided in the supplementary material.
He then added these examples to the prompt and re-ran the model.
Under this limited intervention, the previously incorrect cases matching this pattern were corrected, and the benchmark accuracy in this run increased from \textbf{81.5\%} to \textbf{85.5\%} on the analyzed set.

\section{Expert Interview}

To assess the usability and effectiveness of \tool, we conducted one-on-one interviews with six experts (\textbf{\emph{P1}}--\textbf{\emph{P6}}) who had not participated in the system design process.
\textbf{\emph{P1}} is an associate professor with long-term research expertise in multimodal AI, deep learning, and large language models for intelligent healthcare.
\textbf{\emph{P2}} is an associate professor focusing on medical AI with six years of research experience.
\textbf{\emph{P3}} is a Ph.D. student with three years of experience in medical large language models.
\textbf{\emph{P4}} is a senior Ph.D. student with four years of research experience in multimodal AI.
\textbf{\emph{P5}} is a student in computer science with two years of experience in medical large language model analysis.
\textbf{\emph{P6}} is a senior undergraduate student with two years of research experience in LLM fine-tuning.
The evaluation was conducted via an online meeting with screen sharing.
Each session began with a 15-minute introduction to the research background and our system, followed by 10 minutes of free exploration and 20 minutes to complete five designated tasks.
Participants then spent 5 minutes organizing their observations, followed by a 10-minute semi-structured interview on the workflow, visual design, interactions, and possible improvements.
The task list and interview questionnaire are provided in the supplementary material.
{The interview was covered by the same IRB approval.}

\textbf{System Workflow.}
Before using \tool, most participants relied on manually reviewing erroneous cases and their reasoning text, while a few also used LLM-based scripts for auxiliary analysis.
Across these workflows, participants found it difficult to identify common characteristics across negative samples, compare erroneous and correct reasoning in a structured manner, and turn scattered observations into actionable debugging insights.
In contrast, all participants agreed that \tool better matches their reasoning debugging logic by organizing the process into a coherent workflow.
They particularly appreciated that the system makes error types visually explicit and helps them quickly locate recurrent problems across cases.
\textbf{\emph{P5}} and \textbf{\emph{P6}} emphasized that the knowledge graph grounding was especially helpful because it provided external medical knowledge support during inspection, while \textbf{\emph{P3}} noted that the regional clustering and path comparison helped reveal likely causes behind frequent errors.
\textbf{\emph{P1}} and \textbf{\emph{P2}} further remarked that the system reduced repetitive manual inspection and improved the overall efficiency of reasoning debugging.
\textbf{\emph{P3}} and \textbf{\emph{P4}} additionally noted that, if connected to a broader external medical knowledge graph in the future, the same workflow could be extended to more specialized medical subdomains.
Overall, participants agreed that the system supports the identification of recurrent error patterns across multiple cases and offers more structured evidence than their previous workflows.

\textbf{Visual Designs and Interactions.}
All participants agreed that the visual designs were generally well matched to the analysis tasks.
They found the coordinated views effective for moving between aggregate summaries and detailed case evidence, and considered the interactions sufficient for their debugging needs.
Among the individual components, the \textit{Error View} was the most frequently mentioned as particularly useful because it directly contrasts erroneous reasoning patterns with their corresponding reference reasoning alternatives.
\textbf{\emph{P4}} and \textbf{\emph{P5}} also positively commented on the glyph design in the \textit{Path View} and the heatmap-based regional summary in the \textit{Projection View}, noting that these designs allowed them to quickly identify which entities were most problematic and to grasp the corresponding reasoning trajectories.
Overall, the interview results suggest that the visual encodings and interactions supported both high-level pattern discovery and case-level verification in a coherent manner.


\revise{
\textbf{Critical Feedback and Suggestions.}
\cll{SR.2, R2.7, R3.2}
\textbf{\emph{P6}} noted that when many entities are displayed in the \textit{Path View}, it can be difficult to quickly locate a target node.
\textbf{\emph{P1}} also mentioned that first-time users would benefit from a more comprehensive onboarding tutorial to better understand the system's concepts, visual encodings, and interactions.
\textbf{\emph{P3--P5}} suggested that expanding the coverage of the external knowledge graph could further improve applicability to broader medical scenarios.
For example, \textbf{\emph{P5}} noted that extending the KG with surgery-specific knowledge, such as anatomical structures and surgical procedures, could better support debugging in surgery-related cases.
\textbf{\emph{P2}} and \textbf{\emph{P5}} also observed that clustered medical concepts may contain terminology that is difficult for LLM developers to interpret, and suggested simplifying the displayed language or linking concepts to external references.
}

\section{Discussion}

In this section, we reflect on the design implications of \tool for knowledge-grounded reasoning debugging, including the role of external knowledge and structured error abstractions, and discuss its current limitations and opportunities for future work.

\subsection{Design Implications}
\textbf{Leveraging External Knowledge for Debugging.}
When developers lack domain expertise, we consider incorporating external medical knowledge, such as clinical guidelines, medical literature, and expert-authored references for reasoning debugging.
They provide rich contextual information, but their largely unstructured form makes it difficult to align model reasoning traces and compare across cases.
KGs offer a structured alternative for alignment, path construction, and structural comparison. 
We therefore adopt a Bio-KG, as it preserves diagnostically important entities and relations in a form well suited for systematic reasoning analysis \cite{su2023ibkh,cui2025kghealthcare,abusalih2023healthcarekg}.
\revise{
Following prior KG-grounded path generation strategies~\cite{wu2025medreason}, we extract paths as clinically plausible structural references, as they are derived from biomedical entities and typed relations in the Bio-KG. 
LLM pruning removes case-irrelevant paths without adding or modifying KG edges. 
These KG-supported references help identify reasoning steps that lack biomedical support.
However, missing errors may be under-detected when valid alternative diagnostic paths are absent, while relation and branch errors may be affected by incomplete or outdated KG relations. 
Future work could improve reference reliability with alternative paths, updated knowledge sources, and expert verification for clinically sensitive cases~\cite{hoyt2019recuration,abusalih2023healthcarekg}.
}

\textbf{Designing Reasoning Error Types Based on the Knowledge Representation.}
Rather than starting from a generic taxonomy of model errors, we derive error abstractions directly from the structural units of the underlying knowledge graph. When model reasoning is represented as nodes and directed edges, each reasoning step can be examined from three complementary aspects: the start node, the connecting edge, and the end node.
This structural perspective makes the resulting error types both operational and tightly aligned with the objects actually being analyzed. Each error category corresponds to a distinct failure mode within a single reasoning step, reducing conceptual overlap while maintaining sufficient coverage for downstream analysis. More generally, when model behavior is analyzed as a structured process rather than free-form text, error taxonomies can be systematically grounded in the failure modes of the chosen representation, enabling more coherent visual encodings and debugging workflows.

\textbf{Generalizability and scalability.}
Although \tool is developed for medical diagnosis, its workflow is not limited to this application. The workflow structures model reasoning as paths, grounds them in external domain knowledge, derives comparable references and reasoning errors, and supports visual analysis of recurrent error patterns.
The workflow can extend to other reasoning-intensive domains when model outputs can be represented as structured reasoning units and an external reference space is available for grounding and comparison.
\revise{
Nevertheless, the \textit{Path View} faces visual scalability limits as datasets grow, since more reasoning paths and circular glyphs may be displayed, making local structures harder to read.
}
\revise{
We partly mitigate this issue by filtering nodes according to error intensity and supporting zooming for focused inspection.
Future work could further reduce visual clutter by aggregating paths based on semantic groups or KG regions~\cite{mccray2001aggregating}.
}

\subsection{Limitations and Future Work}
Our work has several directions that deserve further study.
1) The system depends on entity extraction and KG alignment.
Although we adopt the alignment strategy following prior medical reasoning work~\cite{wu2025medreason}, ambiguous biomedical mentions may still introduce grounding errors. Future work can incorporate uncertainty-aware alignment or support user revision of low-confidence matches.
\revise{
2) Our current setting focuses on diagnosis-oriented QA datasets, which  provide relatively structured inputs and ground-truth diagnoses. Future work could extend the framework to more realistic diagnostic data, such as unstructured clinical notes, longitudinal records, and multimodal evidence.
%
}
3) The current system is demonstrated on open-source LLMs. 
While the proposed diagnosis framework is model-agnostic,
future work can extend the evaluation to proprietary LLMs and compare whether different model families exhibit distinct reasoning-level failure modes.
4) The current system
supports reasoning-level diagnosis rather than automatic or immediate model refinement.
It identifies where and how model reasoning deviates from reference KG paths, but the interpretation and use of these diagnoses currently still require users' expertise and manual follow-up.
Future work should investigate how the diagnosed error patterns can be translated into concrete improvement strategies under different model-access settings.
For closed-source models, the identified errors may guide improvement strategies such as injecting KG-aware constraints into prompts, steering retrieval toward evidence that preserves entity-level reachability to candidate diagnoses, and deprioritizing passages that reintroduce known error transitions during evidence reranking.
For open-source models, where training or adaptation is possible, the identified patterns could additionally guide targeted data augmentation, supervised fine-tuning, or reinforcement learning, for instance by converting error cases into corrective training examples via preference optimization~\cite{rafailov2023direct}, incorporating KG-aware constraints into instruction-tuning data~\cite{wu2025medreason}, or shaping reward signals from KG reachability to penalize unsupported entity transitions~\cite{yan2025rlkgf}.

\section{Conclusion}

In this work, we presented \tool, a visual analytics system to help medical LLM developers debug medical diagnostic reasoning with external knowledge.
The system constructs structured reasoning paths aligned with a Bio-KG, and identifies three types of reasoning errors.
Interactive visualizations support structured comparison between observed and reference reasoning paths, while the Sankey-based error analysis enables users to identify recurrent error patterns in reasoning.  
Two case studies and the expert interview demonstrated the system's potential to help users uncover recurrent reasoning errors, interpret their medical implications, and motivate the refinement hypotheses, suggesting its value for practical reasoning-level debugging.


\acknowledgments{%
We thank the reviewers for their insightful feedback.
All human-participant studies reported in this paper were approved by the Human and Artefacts Research Ethics Committee of The Hong Kong University of Science and Technology (Application No.~HREP-2025-0204).
Informed consent was obtained from all participants.
This work is partially supported by Hong Kong Research Grants Council under the Areas of Excellence Scheme grant AoE/P-601/23-N.
}

\bibliographystyle{abbrv-doi-hyperref}
\bibliography{template}

\end{document}